%% file: main.tex
\documentclass[letterpaper]{article} % DO NOT CHANGE THIS

\ifdefined\aaaianonymous
    \usepackage[submission]{aaai2026}  % Anonymous submission version
\else
    \usepackage{aaai2026}              % Camera-ready version
\fi

\usepackage{times}        % DO NOT CHANGE THIS
\usepackage{helvet}       % DO NOT CHANGE THIS
\usepackage{courier}      % DO NOT CHANGE THIS
\usepackage[hyphens]{url} % DO NOT CHANGE THIS
\usepackage{graphicx}     % DO NOT CHANGE THIS
\usepackage{natbib}       % DO NOT CHANGE THIS, NO OPTIONS
\usepackage{caption}      % DO NOT CHANGE THIS, NO OPTIONS
\usepackage{booktabs}
\usepackage{multirow}
\usepackage{amsmath}
\usepackage{amssymb}
\usepackage{xcolor}
\usepackage{algorithm}
\usepackage{algorithmic}

\title{Encoded but Disconnected:
Decomposing Vision--Language Model Failures
under a Patching Null}

\author{
    Genpei Zhang
}
\affiliations{
    University of Wisconsin--Madison\\
    genpei.zhang@wisc.edu
}

\begin{document}

\maketitle

\begin{abstract}
Across three vision--language model architectures (LLaVA-1.5-7B,
Qwen2.5-VL-7B, InternVL3-8B), we report a universal negative finding for
mid-layer interpretability. On POPE---the benchmark common to all
three---the mid layers encode the ground-truth answer in $68$--$91\%$ of
errors, yet this signal is \emph{not} causally active for the final
prediction: residual-stream patching yields $0\%$ non-trivial flip at the
layer level on all three architectures, and on two of three at the
per-head level (Qwen: $0/12{,}600$ patched forwards). The lone exception,
InternVL3 layer-20 head-2, is a non-vocab, self-attending head whose
effect is localized to that specific head ($p<10^{-4}$). Despite the null,
the errors separate operationally into three failure modes---Perception
Failure, Encoded-but-Disconnected, Prior-Override---learnable above
$60\%$ on all three architectures, and the architecture's prior direction
predicts which of two interventions elicits a category-specific response.
We report these mitigation effects under oracle labels as evidence the
categories are mechanistically real, not as a deployable method.
\end{abstract}

% \begin{links}
%     \link{Code}{https://example.com/code}
% \end{links}

% --- Sections (per Step 5 section-by-section drafting) ---
\input{sections/01_intro}
\input{sections/02_related}
\input{sections/03_method}
\input{sections/04_experiments}
\input{sections/05_mitigation}
\input{sections/06_discussion}

% aaai2026.sty already sets \bibliographystyle internally — do not override.
\bibliography{references}

\clearpage
% Step 7 L21 — unified \input for reading; strip for final submission if AAAI
% requires separate supplementary upload.
\input{sections/99_appendix}

\end{document}

%% file: sections/01_intro.tex
\section{Introduction}
\label{sec:intro}

% Pattern A — Schaeffer "alternative reading" opener (per reference_papers_diagnostic.md §5).
% Target: 600-800 words. 3 paragraphs + in-prose contribution sentence.

Recent work characterizes vision--language model (VLM) failures via aggregate
hallucination rates on benchmarks such as POPE~\citep{li2023pope}, MMVP~\citep{tong2024mmvp},
and HallusionBench~\citep{guan2024hallusionbench}, and proposes decoding-time
mitigations that apply uniform interventions across samples: visual contrastive
decoding~\citep{leng2024vcd}, over-trust penalties~\citep{huang2024opera},
multi-modal grounding~\citep{favero2024m3id}, and layer-contrast
methods~\citep{chuang2024dola}. Here we present an \emph{alternative reading}:
when probed at the per-error level via mid-layer representations, a single
benchmark error is not a single phenomenon. It decomposes into mechanistically
distinct failure modes whose mid-layer correlates point in opposite directions,
and which a uniform mitigation can only inconsistently address.

Concretely, we ask three questions of every error a VLM makes on a binary or
discriminative prompt. (1) \emph{Did the vision encoder transmit usable
evidence for the ground-truth answer?}, a probe-style test on mid-layer
features. (2) \emph{Does the model already encode the ground-truth answer
at the mid layers yet emit the wrong final token?}, the Encoded Answer
Signal (EAS), operationalized via raw logit
lens~\citep{nostalgebraist2020logitlens,belrose2023tunedlens}. (3)
\emph{When the model emits the wrong answer, does that wrong answer match
the direction of its training prior?}, an empirical-prior test. The three
answers partition each error into Perception Failure, Encoded-but-Disconnected,
or Prior-Override. The decomposition is operational, not interpretive: each
class is defined by observable mid-layer and final-layer behavior on a
single forward pass.

We carry the decomposition through three architectures
(LLaVA-1.5-7B~\citep{liu2024llava}, Qwen2.5-VL-7B~\citep{bai2025qwen25vl}, and
InternVL3-8B~\citep{zhu2025internvl3}) and two benchmark genres (POPE binary
hallucination and HallusionBench/MMVP visual discrimination). Three findings
emerge. First, the EAS phenomenon is large and architecture-universal: on
POPE, the benchmark common to all three, $68$--$91\%$ of errors carry a
mid-layer top-1 hit on the ground-truth answer despite a wrong final-layer
top-1. Second, this encoded signal is \emph{not} causally active for the
final token: residual-stream patching at the encoding layers fails to flip
predictions on strict cross-instance pairs at the layer level on all three
architectures, and at the per-head level on two of three (the Qwen sweep
alone covers $12{,}600$ patched forwards with zero flips). On InternVL3,
the per-head signal concentrates at a single non-vocab, self-attending
head---a behavior incompatible with a direct decoding-level encoder. Third, the
population mix of failure classes differs by prior direction in a way that
predicts which intervention elicits a category-specific response:
under-affirmative models (Qwen, InternVL3) respond to prompt-debias on the
Prior-Override class ($\sim$17--20 percentage points), while the
over-affirmative model (LLaVA) responds to mid-layer Re-Promote on the
Encoded-but-Disconnected class. These effects are measured under oracle
category labels.

\paragraph{Contributions}
We contribute, in prose: (i) a three-way \emph{operational} decomposition of
VLM failures, defined from a single forward pass and separable from
continuous mid-layer features above the $60\%$ threshold on all three
architectures (on LLaVA, the features add $\sim$20 points over a
prediction-direction baseline); (ii) a paper-grade negative finding that
mid-layer encoding of the ground-truth answer is not causally active for
the final prediction under standard activation patching---universal at the
layer level on three architectures and the per-head level on two---which
challenges the assumption that lens-accessible information is causally
relevant; (iii) the discovery of an architecture-specific causal head
(InternVL3 layer 20, head 2): a $(\text{layer},\text{head})$-localized
effect ($p<10^{-4}$ vs.\ a different head) that is not a yes/no vocab
encoder and is self-attending; and (iv) a prior-direction-aware mitigation
framework in which each architecture's dominant failure mode admits a
different per-category intervention, demonstrated under oracle category
labels. We release code, model SHAs, and the per-sample classifications
used in every figure.

Our results elevate vision--language mechanistic interpretability from
population-level diagnostic to per-sample, prior-aware routing, and show
that the question ``where does the signal go?'' has a different
mechanistic answer on each architecture even when the phenomenology is
universal. Related Work situates this against recent
three-way taxonomy and mid-layer promotion lines of work;
Method defines the EAS and decomposition operationally;
Experiments presents the three-architecture evidence
including the patching null and the L20--H2 mechanism;
Per-Category Intervention reports the per-category mitigation results; and
Discussion closes with what we do not claim and what
remains for future work.

%% file: sections/02_related.tex
\section{Related Work}
\label{sec:related}

% Target: 500-700 words. Three thematic paragraphs (workflow doc Hard rule for §2).
% Differentiators against R1a (Anatomy of a Lie) and R10 (VISTA) must be explicit.

\paragraph{VLM hallucination phenomenology and benchmarks}
Object hallucination and visual misperception in vision--language models have
been documented across multiple benchmark families. POPE~\citep{li2023pope}
uses binary yes/no prompts about object presence; MMVP~\citep{tong2024mmvp}
constructs CLIP-similar image pairs to expose visual-feature shortcuts;
HallusionBench~\citep{guan2024hallusionbench} curates entangled language and
visual illusions; MME~\citep{fu2023mme} aggregates fourteen perception and
cognition subtests; MMStar~\citep{chen2024mmstar} re-audits prior benchmarks
for visual necessity. Each surfaces an aggregate failure rate. None decomposes
a single benchmark error into its mechanistic origin, and recent diagnostic
work~\citep{caos2025} measures hallucination via downstream metrics rather
than per-error internal probes. Our diagnostic operates one level below: it
classifies the error from the same forward pass that produces it.

\paragraph{Mechanistic interpretability for vision--language models}
The logit lens~\citep{nostalgebraist2020logitlens} and its trained
variant~\citep{belrose2023tunedlens} read mid-layer hidden states as if they
were final-layer predictions, and have been extended to multimodal models in
work that locates ``multimodal neurons'' inside text-only
LLMs~\citep{schwettmann2023multimodal}. Activation
patching~\citep{meng2022rome, wang2023ioi} causally tests whether
a localized representation contributes to a target prediction.
Concurrent with our work, a three-class decomposition of VLM
hallucination~\citep{anatomyofalie2025} names perceptual instability,
logical-causal failure, and decisional ambiguity at the population level.
We differ in three ways: we (i) carry the decomposition across three
architectures, (ii) make the per-sample classifier explicit and define the
categories from a single forward pass, and (iii) route to a class-specific
mitigation. A concurrent tri-layer framework~\citep{tosee2026} similarly
separates perception, dependency, and alignment failures, but does not
test the causal status of the mid-layer signal. Crucially, we add a
paper-grade negative finding on activation
patching---a universal null at the layer level on three architectures plus
a per-head null on two---that directly tests, and refines, the implicit
assumption in this line of work that lens-accessible information is
causally relevant.

\paragraph{Decoding-time mitigation and the gap our work fills}
Existing decoding-time mitigations apply uniform interventions across samples.
VCD~\citep{leng2024vcd} subtracts logits computed under a perturbed visual
condition; OPERA~\citep{huang2024opera} penalizes attention over-trust and
back-tracks generation; M3ID~\citep{favero2024m3id} re-anchors decoding to
visual evidence; DoLa~\citep{chuang2024dola} contrasts an early and a late
layer. A recent line of work, VISTA~\citep{vista2025}, combines an
activation-space visual steering vector with early-layer logit promotion
during decoding, and reports a $\sim$40\% hallucination reduction on
open-ended generation. Two distinctions from our Re-Promote component:
VISTA applies its promotion uniformly across all samples, while we
(i) promote from the sweet-zone mid-layer specifically (chosen
per-architecture via the EAS phenomenon), and (ii) gate the promotion on
the predicted failure category, so it does not target the prior-override
class where the mid-layer disagrees in the wrong direction. On the
over-affirmative LLaVA architecture, under oracle category labels the gated
promotion separates the encoded-but-disconnected and prior-override flip
rates by a large category-conditional margin ($85.9\%$ vs.\ $0\%$) that a
uniform intervention cannot, by construction, produce.

In sum, prior work has characterized VLM failure phenomenologically, probed
mid-layer representations diagnostically, and proposed uniform decoding
interventions. The piece we contribute is mechanism-aware routing: a single
forward pass classifies the failure mode, and the classification predicts
which of two prior-direction--appropriate interventions to apply. The
classification itself is grounded in a three-architecture patching study
that includes the universal null and the L20--H2 exception, both
documented in Experiments. Operational distinctions
between our intervention and the five most-relevant baselines (VCD,
OPERA, M3ID, DoLa, VISTA) are spelled out in Appendix~\ref{app:baselines}.

%% file: sections/03_method.tex
\section{Method}
\label{sec:method}

% Target: 600-900 words. Lead with a question (workflow doc Hard rule #10).
% Push hyperparams to Appendix\ref{app:main}.

\subsection{The Encoded Answer Signal (EAS)}

What does it mean for a vision--language model to ``know'' the answer it does
not emit? We make this precise via the raw logit lens. Let $h^{(\ell)}_t$
denote the residual-stream hidden state at layer $\ell \in \{1, \dots, L\}$,
position $t$ (we use the last-text position throughout). Applying the model's
own final layernorm $\phi$ and unembedding matrix $W_U$ to $h^{(\ell)}_t$
yields a mid-layer prediction distribution
$\pi^{(\ell)}_t = \mathrm{softmax}(W_U \phi(h^{(\ell)}_t))$. We say the
mid-layer \emph{encodes} the ground-truth answer token $y^\star$ when there
exists at least one $\ell \in [\ell_{\min}, \ell_{\max}]$ (the
architecture-specific ``sweet zone,'' defined below) such that
$\arg\max \pi^{(\ell)}_t = y^\star$. An error sample $(x, y^\star)$ exhibits
the EAS phenomenon when this condition holds at mid layer but the final
prediction $\arg\max \pi^{(L)}_t \neq y^\star$. EAS is thus a per-sample
binary property of a single forward pass. The aggregate EAS top-1 rate over
an error set is formalized in Appendix~\ref{app:metrics}.

The sweet zone is the layer window over which we scan for a mid-layer
top-1 hit, chosen by maximizing aggregate top-1 hit rate on a held-out
subset of correctly-answered samples (not the test errors): $[12,24]$ for
LLaVA-1.5-7B, $[10,21]$ for Qwen2.5-VL-7B, $[6,26]$ for InternVL3-8B. The
hit rate peaks at a narrower band within each window ($12$--$13$,
$10$--$17$, $18$--$21$; roughly $\tfrac13$, $\tfrac12$, $\tfrac23$ relative
depth given $32/28/28$ layers), which we use for relative-depth statements;
the full window defines the EAS condition. Both are tabulated in
Appendix~\ref{app:main}. The deepening peak with architecture is a finding
discussed in Experiments.

We deliberately use the raw lens---no learned adapter---throughout this paper.
A tuned-lens variant~\citep{belrose2023tunedlens} was trained as a separate
ablation but is not load-bearing: on LLaVA the raw lens already reaches a
$98.5\%$ top-5 hit rate at L19 on correct samples
(Appendix~\ref{app:metrics}); for Qwen and InternVL3 we report only the
error-set EAS rate, as no correct-sample top-5 audit was collected. The
diagnostic does not depend on a trained probe.

\subsection{Three-Way Decomposition of Errors}

Each error is assigned to one of three categories from two binary
observations: (a) the EAS condition above, and (b) whether the emitted
token $\hat{y}$ lies in the \emph{prior-direction} set $T_{\text{prior}}$,
determined empirically per architecture---``yes'' tokens for
over-affirmative models, ``no'' for under-affirmative ones
(Table~\ref{tab:models}, Appendix~\ref{app:main}):

\begin{itemize}
  \item \textbf{Prior-Override (PO)}: $\hat{y} \in T_{\text{prior}}$. The model
        emitted in the direction of its training prior; whether or not the
        mid-layer also held that direction, the prior won the final decision.
  \item \textbf{Encoded-but-Disconnected (EBD)}: $\hat{y} \notin T_{\text{prior}}$
        and EAS condition is true. The mid-layer encoded $y^\star$, but the
        final layer collapsed to the opposite-of-prior answer without
        propagating the encoded signal.
  \item \textbf{Perception Failure (PF)}: $\hat{y} \notin T_{\text{prior}}$ and
        EAS condition is false. Neither the mid-layer nor the final layer
        carries the ground-truth answer; the failure is upstream of the LM
        backbone.
\end{itemize}

The three classes are mutually exclusive and exhaustive over errors. Note that
the labels are defined operationally, not interpretively: an ``EBD'' label
does not assert that the mid-layer ``knows the answer''; it asserts only the
two observable conditions, and we test the causal interpretation explicitly
in Experiments.

\subsection{Activation Patching Protocol}

To test whether the mid-layer encoded signal is causally active for the
final prediction, we apply standard cross-instance activation patching.
Given an error sample $(x_e, y_e^\star)$ and a ``donor'' $(x_d, y_d^\star)$
on which the model is correct, we patch the residual stream
$h^{(\ell)}_t$ of $x_e$ with that of $x_d$ and re-run from $\ell$ onward; a
success is a patched prediction that flips to $y_e^\star$. We restrict to
strict pairs ($y_e^\star \neq y_d^\star$ at the label level), for which a
flip can only reflect a causal effect of the patched representation---this
excludes the ``answer-transplant'' artifact where a correct-label donor
simply imprints its own label.

We sweep the patching $\ell$ over the sweet zone and the patching position
over $\{\text{BOS}, \text{visual-last}, \text{last-text}\}$ for the
layer-level protocol, and additionally over individual attention heads
$h \in \{0, \dots, H-1\}$ for the per-head protocol. Per-architecture sweep
counts (the Qwen per-head sweep alone is $12{,}600$ forwards; layer-level
and other per-head counts are smaller) are reported with their denominators
in Experiments.

\subsection{Per-Sample Classifier and Per-Class Mitigation}

Algorithm~\ref{alg:pipeline} summarizes the full per-sample pipeline: a
single forward pass produces the 24-dimensional feature vector and the
operational category label, the classifier predicts category on unseen
samples, and the category-specific intervention is applied.

\begin{algorithm}[!t]
\caption{Per-sample failure-mode routing. Lines using $T_{y^\star}$
(ground-truth) are diagnostic-only; a deployed router drops the
correctness gate on line~3 and routes all samples via $C$.}
\label{alg:pipeline}
\begin{algorithmic}[1]
\REQUIRE image $x$, question $q$, VLM $M$ with sweet-zone $[\ell_{\min}, \ell_{\max}]$,
        prior-direction set $T_{\text{prior}}$, classifier $C$ (pre-trained),
        interventions $\mathcal{I}_{\text{PO}}$ (prompt-debias) and
        $\mathcal{I}_{\text{EBD}}$ (mid-layer Re-Promote)
\STATE $(\pi^{(1)}, \dots, \pi^{(L)}, z^{(L)}) \leftarrow M(x, q)$  \hfill\emph{// one forward pass}
\STATE $\hat{y} \leftarrow \arg\max \pi^{(L)}$
\IF{$\hat{y} \in T_{y^\star}$ (correct)}  \STATE \textbf{return} $\hat{y}$  \ENDIF
\STATE $\text{eas\_hit} \leftarrow \exists \ell \in [\ell_{\min},\ell_{\max}]:
        \arg\max \pi^{(\ell)} \in T_{y^\star}$
\STATE $\text{cat} \leftarrow$ PO if $\hat{y} \in T_{\text{prior}}$
        else (EBD if eas\_hit else PF)
\STATE $\mathbf{f} \leftarrow$ \texttt{features}($\pi^{(\ell_{\min} \ldots \ell_{\max})}, \pi^{(L)}$)
        \hfill\emph{// 24-d}
\STATE $\hat{c} \leftarrow C(\mathbf{f})$  \hfill\emph{// at inference; cat for training}
\IF{$\hat{c} = \text{PO}$}     \STATE \textbf{return} $\arg\max M(\mathcal{I}_{\text{PO}}(x, q))$
\ELSIF{$\hat{c} = \text{EBD}$} \STATE \textbf{return} $\arg\max
        (z^{(L)} + \alpha \cdot W_U \phi(h^{(\ell^\star)}))$
\ELSE \STATE \textbf{return} $\hat{y}$  \hfill\emph{// PF: no helpful intervention}
\ENDIF
\end{algorithmic}
\end{algorithm}

Given a forward pass, the three-class label and a 24-dimensional feature
vector (per-layer top-1/2/3 logits and margins, attention-mass summaries,
final-layer top-1/2 logits and margin; full list in
Appendix~\ref{app:main}) are extracted within the same pass, with no extra
inference. The features exclude the token-identity and EAS-hit indicators
that define the labels, so the classifier cannot trivially reconstruct its
target (Appendix~\ref{app:clf}). A gradient-boosted decision tree (GBDT,
$n=100$, depth 3) and a logistic-regression variant are trained per
architecture under 5-fold cross-validation. The classifier serves as the
routing layer of the
mitigation pipeline: each predicted category is mapped to the
prior-direction--appropriate intervention defined in Per-Category
Intervention (prompt-debias for PO on No-prior models, mid-layer Re-Promote
for EBD on the Yes-prior model). Training data are the labeled error samples
themselves, weakly supervised by the operational definition above; no human
annotation is used.

%% file: sections/04_experiments.tex
\section{Experiments}
\label{sec:experiments}

% Target: 1100-1500 words. Open with question (Hard rule #10). Lead each
% subsection with a CLAIM, support with numbers, interpret. Tables to push to
% appendix; main paper holds the headline numbers.

If aggregate accuracy hides a mechanistically structured failure, the
diagnostic should expose it: each failure category should appear in
non-trivial proportion, the labels should be separable from internal
features, and---the question we test most sharply---the mid-layer signal
that defines Encoded-but-Disconnected should be \emph{causally}, not merely
correlationally, tied to the final prediction. We organize the
three-architecture evidence around four tests: phenomenology, classifier
separability, the causal null, and one architecture-specific exception.
Table~\ref{tab:master} previews all four across the three architectures.

\begin{table*}[!t]
\centering
\small
\begin{tabular}{lccccccc}
\toprule
 & & \multicolumn{2}{c}{Phenomenology} & \multicolumn{2}{c}{Patching null} & \multicolumn{2}{c}{Diagnostic}\\
\cmidrule(lr){3-4}\cmidrule(lr){5-6}\cmidrule(lr){7-8}
Architecture & Family & EAS (POPE) & Peak band & Layer level & Per-head & CV acc & Target flip\\
\midrule
LLaVA-1.5-7B  & CLIP + LLaMA-2     & 74.5\% & L12--L13 ($\tfrac{1}{3}$) & $0 / 50$ & $1/32$ (noise) & $82.5 \pm 3.8$ & EBD $85.9$\%\\
Qwen2.5-VL-7B & SigLIP + Qwen2     & 91.0\% & L10--L17 ($\tfrac{1}{2}$) & $0 / 44$ & $0 / 12{,}600$ & $68.0 \pm 3.9$ & PO $19.6$\%\\
InternVL3-8B  & InternViT + Qwen2  & 68.5\% & L18--L21 ($\tfrac{2}{3}$) & $1 / 50$ & \textbf{L20 H2: 26\%} & $79.7 \pm 3.2$ & PO $21.3$\%\\
\bottomrule
\end{tabular}
\vspace{-2mm}
\caption{Three-architecture summary of all main-paper results.
\textbf{EAS (POPE)}: mid-layer top-1 hit rate on POPE errors. \textbf{Peak
band}: layers of strongest signal (relative depth). \textbf{Patching
null}: per-pair non-trivial flip rate at the layer and per-head levels;
InternVL3 L20 H2 is the only tuple rejecting the null. \textbf{CV acc}:
5-fold accuracy of the 24-feature classifier on PF/EBD/PO (logistic
regression). \textbf{Target flip}: flip rate of the targeted class under
oracle labels. Details in the appendices.}
\label{tab:master}
\vspace{-3mm}
\end{table*}

\subsection{Setup}

Three architectures are evaluated end-to-end: LLaVA-1.5-7B
(CLIP-ViT-L + LLaMA-2)~\citep{liu2024llava, liu2023llava}, Qwen2.5-VL-7B
(SigLIP + Qwen2)~\citep{bai2025qwen25vl}, and InternVL3-8B (InternViT +
Qwen2)~\citep{zhu2025internvl3}. Inference uses fp16 (LLaVA, Qwen) or bf16
(InternVL3); we record model SHAs in Appendix~\ref{app:main}. Evaluation
benchmarks are POPE adversarial split for binary hallucination, plus
MMVP (LLaVA only) and HallusionBench for visual-discrimination tasks.
Per-benchmark prompt formats and sample counts are summarized in
Appendix~\ref{app:datasets}.
Errors are defined as a strict first-token mismatch against the
ground-truth answer's token set. Per-error category labels follow the
operational definition in Method; all extraction code
operates on a single forward pass per sample.

\subsection{Phenomenology: EAS is large and architecture-universal}

\textbf{Claim 1.} The Encoded Answer Signal is present in the majority of
errors on all three architectures, and the strongest signal sits at a
different relative depth on each.

On POPE, the benchmark common to all three, the mid-layer top-1 hit rate
on errors is $74.5\%$ (LLaVA), $91.0\%$ (Qwen), and $68.5\%$ (InternVL3).
The phenomenon is not confined to binary prompts: pooled over each
architecture's full benchmark set it remains substantial ($52\%$ LLaVA
over POPE+MMVP, $81\%$ Qwen and $68\%$ InternVL3 over POPE+HallusionBench;
the lower LLaVA pool reflects MMVP's harder $29.5\%$ rate). Because the
pools differ across architectures, we use POPE for cross-architecture
comparison and report the rest in Appendix~\ref{app:metrics}. EAS is thus
not a peculiarity of any one backbone or vision encoder.

The peak layer shifts with architecture: LLaVA $12$--$13$ ($\sim\tfrac13$
depth), Qwen $10$--$17$ ($\sim\tfrac12$), InternVL3 $18$--$21$
($\sim\tfrac23$). The transition is sharpest on LLaVA, where the top-5 hit
rate on correct samples jumps from $57.7\%$ at L18 to $98.5\%$ at L19;
analogous but wider transitions appear on the other two. Architecture
differs in \emph{where} the encoded answer crystallizes, not whether it
does.

\subsection{Classifier learnability and bimodal POPE}

\textbf{Claim 2.} The three categories are separable from a 24-feature
classifier above the $60\%$ threshold on all three architectures---though
on the prior-dominated models a prediction-direction baseline is already
strong---and the POPE separation is bimodal.

On the PF/EBD/PO labels, a gradient-boosted tree (GBDT, $n=100$, depth 3)
and a logistic-regression (LR) variant reach 5-fold CV accuracy of
$82.5\%\pm3.8$ (LLaVA), $68.0\%\pm3.9$ (Qwen), and $79.7\%\pm3.2$
(InternVL3) under LR, with GBDT comparable (Appendix~\ref{app:clf}). The
features exclude the indicators that define the labels, so this is not
label reconstruction; a $2$-bit oracle of those indicators reaches $100\%$,
the continuous features do not. The gain over a prediction-direction
baseline is large only on LLaVA ($62.5\to82.5$), where the categories are
\emph{not} given by prediction direction; on the PO-dominated Qwen and
InternVL3 the direction bit alone is already $93$--$94\%$, so we treat
learnability as a supporting check, not a headline.

\begin{figure*}[!t]
  \centering
  \includegraphics[width=0.9\textwidth]{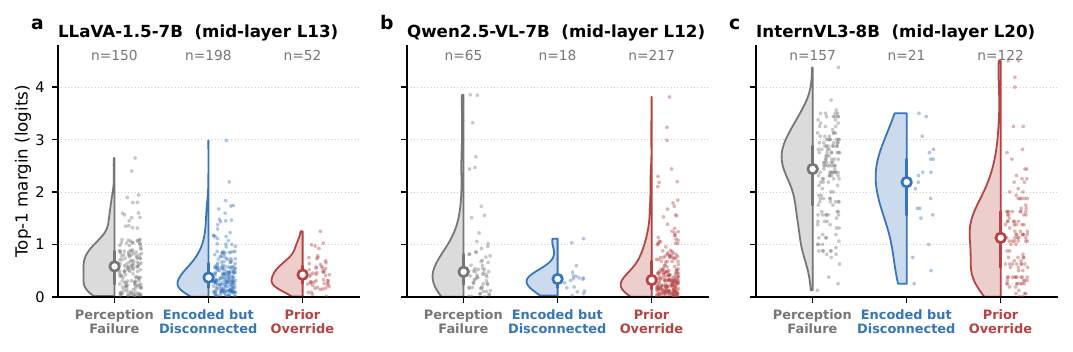}
  \vspace{-3mm}
  \caption{Per-category mid-layer top-1 margin distributions across the
  three architectures (LLaVA L13, Qwen L12, InternVL3 L20). The three
  failure categories separate by margin shape---broad-tailed (PF), low
  (EBD), tight near zero (PO)---which is the signal the 24-feature
  classifier exploits.}
  \label{fig:catdist}
  \vspace{-3mm}
\end{figure*}

\begin{figure}[!t]
  \centering
  \includegraphics[width=0.95\columnwidth]{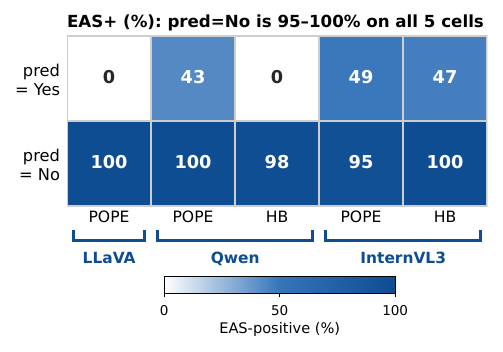}
  \vspace{-4mm}
  \caption{Bimodal POPE failure-mode separation across 5
  model$\times$benchmark cells: pred=No errors are $95$--$100\%$
  EAS-positive (bottom), pred=Yes are $<5\%$ (top). The split is fixed by
  prediction direction; the non-trivial content is that the two halves also
  differ in mid-layer signal.}
  \label{fig:bimodal}
\end{figure}

On binary POPE, the failure mode is bimodal: pred-Yes errors map to
Prior-Override and pred-No errors to Encoded-but-Disconnected, sharply in
$5$ of $5$ model$\times$benchmark cells (Figure~\ref{fig:bimodal}). Two
clarifications. (a) Since the PO label equals ``$\hat{y}\in
T_{\text{prior}}$,'' which on binary POPE \emph{is} the prediction
direction, the $100\%$/$100\%$ split restates the definition rather than
discovering it. The non-trivial content is the \emph{joint} mid-layer
pattern: $95$--$100\%$ of the pred-No half is also EAS-positive (mid-layer
top-1 = ``yes''), versus $<\!5\%$ of the pred-Yes half---the two halves
differ in mid-layer signal, not only final token. (b) The three-class
label still earns its keep downstream: it routes to different
interventions (Per-Category Intervention), whereas the binary
direction split would send PF samples to the wrong one.
Figure~\ref{fig:catdist} shows the mid-layer top-1 margin distribution
separating the three categories on all three architectures, distributional
evidence beyond the labels themselves.

\subsection{The causal null: mid-layer encoding is not causally active}

\textbf{Claim 3.} Despite a large EAS phenomenon, mid-layer activation
patching at the encoding layer does \emph{not} flip final predictions on
strict cross-instance pairs, at the layer level on all three architectures.

\begin{figure*}[!t]
  \centering
  \includegraphics[width=0.95\textwidth]{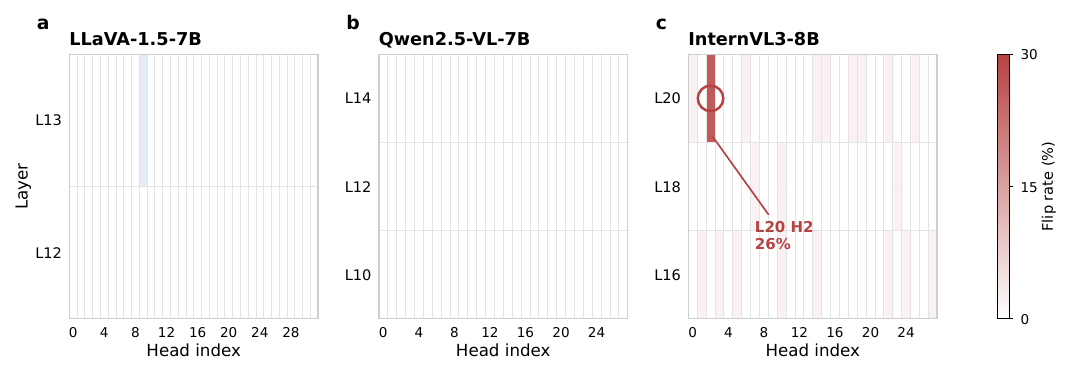}
  \vspace{-5mm}
  \caption{Per-head patching flip-rate heatmaps at the last-text position
  across the three architectures (shared vmax $30\%$). The only cell
  exceeding the layer-level null upper CI is InternVL3 L20 H2 (circled,
  panel c); LLaVA and Qwen are essentially flat at zero.}
  \label{fig:perhead}
\end{figure*}

We perform residual-stream patching at the encoding layers across position
$\{$BOS, visual-last, last-text$\}$ on strict pairs (donor and receiver
have different ground-truth labels). The unit is the \emph{per-pair} flip
rate at each (layer, position) tuple. At the layer level no tuple
flips above noise: LLaVA $0/50$ (Clopper--Pearson upper $95\%$ CI $7.1\%$),
Qwen $0/44$ (upper CI $8.0\%$), InternVL3 a maximum of $1/50$ at any single
tuple (best tuple L2 visual-last, $2\%$; upper CI $10.6\%$; this supersedes
an earlier $0/15$ probe). All are far below the $5\%$ decision threshold. At the per-head granularity the
Qwen sweep ($21$ layers $\times$ $3$ positions $\times$ $32$ heads $\times$
$50$ pairs $= 12{,}600$ attempts) yields zero non-trivial flips at every
tuple. The LLaVA per-head sweep gives $1/32$ at L13 H9, below the Qwen
baseline upper CI, which we treat as noise. Figure~\ref{fig:perhead} shows
the full per-head field; the single InternVL3 hotspot at L20 H2 is the only
cell exceeding the layer-level CI across all three sweeps, and confirms the
methodology detects causal heads when they exist.

We conclude that the mid-layer top-1
encoding of the ground-truth answer in the EBD category, while a robust
phenomenological correlate of failure, is not the locus of the model's
prediction-generation mechanism.

This is a strong negative finding and directly tests the implicit assumption
of work that infers causal role from logit-lens
accessibility~\citep{nostalgebraist2020logitlens, belrose2023tunedlens}.
A natural reading is that the encoded signal sits on a parallel residual
trajectory that the final layer reads only when downstream attention or MLP
state aligns; the patching protocol leaves these downstream dependencies
unaltered, so the encoded signal alone cannot drive the prediction.

\subsection{L20 head 2 on InternVL3: causal but not vocab}

\textbf{Claim 4.} The per-head sweep that confirms the null on LLaVA and
Qwen \emph{rejects} it on InternVL3 at a single (layer, head), but that head
is not a yes/no vocab encoder and attends almost entirely to its own
position---consistent with an architecture-specific aggregation step rather
than a direct decoding pathway.

\begin{figure*}[!t]
  \centering
  \vspace{-2mm}
  \includegraphics[width=0.95\textwidth]{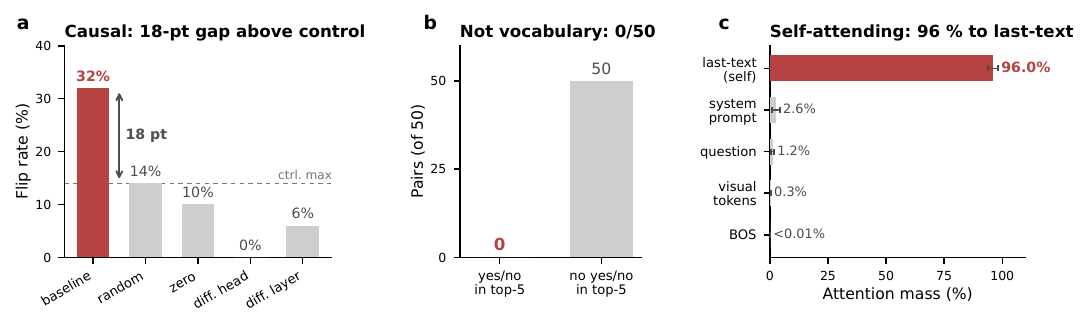}
  \vspace{-6mm}
  \caption{InternVL3 layer 20 head 2 mechanism triplet (the per-head
  exception of Figure~\ref{fig:perhead}). \textbf{(a)} 5-variant sanity
  sweep (baseline vs.\ random, zero, different head, different layer).
  \textbf{(b)} Vocab projection: yes/no tokens absent from the head's
  top-5. \textbf{(c)} Attention: $96\%$ of mass on its own position.
  Causal but not vocab-encoding, and self-attending.}
  \label{fig:l20h2}
  \vspace{-4mm}
\end{figure*}

Per-head patching on InternVL3 reveals one outlier: layer 20 head 2 at the
last-text position produces $26\%$ ($13/50$) non-trivial flips, exceeding
the per-head null elsewhere and the pre-registered $\geq 15\%$ threshold.
An independent 5-variant sanity sweep on a separate $50$-pair sample (hence
a baseline of $32\%$, $16/50$, on that sample) gives: random norm-matched
gaussian $14\%$ ($7/50$), zero vector $10\%$ ($5/50$), different head
(L20 H3) $0\%$ ($0/50$), different layer (L19 H2) $6\%$ ($3/50$). The
effect is localized to the specific (layer, head): Fisher's exact gives
baseline vs.\ different head $p<10^{-4}$ and vs.\ different layer
$p=0.0017$. The comparison against norm-matched random is only borderline
($p=0.056$), so the perturbation magnitude matters alongside its location;
the (layer, head) specificity itself is robust, with the drop to $0\%$ on
the same-layer different-head substitution the single strongest piece of
evidence (Figure~\ref{fig:l20h2}a).

A mechanism probe confirms it is \emph{not} a direct yes/no encoder:
projecting $W_O$ applied to the clean L20 H2 output through the final
layernorm and unembedding lands the yes/no tokens in $0/50$ top-5 lists.
An attention probe shows it places $96.0\%\pm2.1$ of its mass on its own
(last-text) position, $0.3\%$ on visual tokens, $1.2\%$ on the question.
The head is thus causal, non-vocab, and self-attending; a natural
interpretation is that it aggregates residual-stream information within the
last-text position---consistent with the view of late-layer attention
heads in LMs~\citep{geva2023dissecting}---rather than encoding the answer.

This refines rather than contradicts the universal null: no architecture's
mid-layer encoding propagates causally at the layer level, and at the
per-head level two of three stay null while one concentrates a thin causal
pathway at a non-vocab, self-attending head---which also shows the
methodology detects causal heads when they exist.

%% file: sections/05_mitigation.tex
\section{Per-Category Intervention}
\label{sec:mitigation}

% Target: 500-700 words. Pattern C opener (R2/Yuksekgonul "a natural solution is...")
% Honest framing: oracle/conditional results, not deployment-validated end-to-end.

If failures are mechanistically heterogeneous (Method) and that
heterogeneity is separable from internal features (Experiments), a natural
response is to detect the failure type at inference and route each sample
to a category-appropriate intervention. The two interventions act on
different pipeline stages---prompt-debias on the input distribution,
mid-layer Re-Promote on the decoding distribution---so the appropriate one
depends on whether the failure arises from prior pressure (input side) or
disconnected mid-layer encoding (decoding side). We show this on the three
architectures.

\paragraph{Prompt-debias for Prior-Override.}
When the prior overrides correct visual evidence (a pre-decoding failure),
we append a suffix arguing against the model's prior direction. For
under-affirmative models (Qwen, InternVL3): ``Look thoroughly: say yes if
any evidence supports the object's presence; when in doubt, say yes.'' For
the over-affirmative LLaVA the suffix reverses direction. The forward pass
is then repeated.

\paragraph{Mid-layer Re-Promote for Encoded-but-Disconnected.}
When the encoded mid-layer signal fails to reach the final layer, we blend
the mid-layer raw-lens prediction back into the final logits: for a
sweet-zone layer $\ell^\star$ and weight $\alpha$, the prediction becomes
$\arg\max\bigl(z^{(L)}_t + \alpha \cdot W_U \phi(h^{(\ell^\star)}_t)\bigr)$,
with $z^{(L)}_t$ the original final logits. On LLaVA we use $\ell^\star=12$,
$\alpha=1.0$ (selected on a held-out audit; $\alpha$-sensitivity in
Appendix~\ref{app:mitig}).

\textbf{Claim 5.} Under oracle category labels, each architecture admits a
category-targeted intervention with a flip rate well above its
prior-aligned class, and which intervention matches is predicted by the
architecture's prior direction.

For each architecture we label its $200$ POPE errors as PO, EBD, or PF and
apply the prior-direction--appropriate intervention, measuring the
per-category flip rate (the fraction of that category emitting the
ground-truth answer after intervention). Results are in
Table~\ref{tab:mitigation}:

\begin{table}[!t]
\centering
\small
\begin{tabular}{llccc}
\toprule
Arch. & Strategy & flip(PO) & flip(EBD) & flip(PF) \\
\midrule
LLaVA     & Re-Promote & $0.0$          & $\mathbf{85.9}$ & $85.7$ \\
Qwen      & Debias     & $\mathbf{19.6}$ & $0.0$          & $0.0$  \\
InternVL3 & Debias     & $\mathbf{21.3}$ & $3.6$          & $0.0$  \\
\bottomrule
\end{tabular}
\vspace{-3mm}
\caption{Per-category flip rates (\%) under each architecture's matched
intervention; bold marks the targeted class. Counts $k/n$: LLaVA $0/51$,
$122/142$, $6/7$; Qwen $31/158$, $0/24$, $0/18$; InternVL3 $19/89$,
$2/56$, $0/55$. Results are conditional on the operational label, not
end-to-end; LLaVA's PF ($6/7$) shows Re-Promote is not EBD-specific (see
text). Debias and Re-Promote rows use separate runs.}
\label{tab:mitigation}
\end{table}

Two observations follow. First, each architecture's effective strategy
targets its dominant POPE failure mode, which the prior direction predicts:
the over-affirmative LLaVA carries $74\%$ of its POPE errors as EBD, while
the under-affirmative Qwen and InternVL3 carry $79\%$ and $42\%$ as PO
(Table~\ref{tab:classprop}). This is the inter-model signature of our
framework: the \emph{population} composition of failure modes shifts
systematically with prior direction---an over-affirmative model accumulates
errors where the answer is encoded but not emitted (EBD), an
under-affirmative model accumulates errors where the prior overrides the
visual evidence (PO)---which is precisely why the matched intervention
differs by architecture. Second, each intervention is specific to the
prior-\emph{aligned} class rather than to EBD as such: on LLaVA, Re-Promote
leaves PO at $0\%$ but also flips the PF class ($6/7$). PF is essentially
absent on POPE (LLaVA POPE PF$=0$ in the classifier run; the $7$ here come
from the Re-Promote run), so this does not affect the POPE conclusion---but
it shows Re-Promote would lose specificity on tasks where PF is common. We
therefore report the effect as class-conditional, not as evidence of an
EBD-only mechanism.

\paragraph{Asymmetry.}
A secondary pattern emerges across architectures. The under-affirmative
models respond strongly to a positive-tone ``yes when in doubt'' suffix
($17$--$20$ points), whereas the over-affirmative LLaVA responds weakly to
its negative-tone counterpart ($4$ points on PO;
Appendix~\ref{app:mitig}). One reading is instruction-tuning asymmetry---SFT/RLHF
tends to penalize refusal more than affirmation, so a ``yes'' suffix sits
closer to the tuning prior. We flag this only as a hypothesis from the
data: each direction was tested with a single, length-unmatched suffix on
three architectures, so the effect could equally reflect suffix wording;
a phrasing-matched battery is left to future work.

\paragraph{Scope of the evaluation}
The flip rates in Table~\ref{tab:mitigation} are \emph{per-category}: they
condition on a correct predicted label. A deployed system would instead
route by the classifier's prediction, and net benchmark accuracy would
depend jointly on the per-class flip rates and the classifier's
per-class precision/recall (Experiments; above the $60\%$ threshold on all
three architectures). We do not claim a net-accuracy improvement: applying
a single intervention benchmark-wide does not raise net accuracy, since
off-target classes (PO on a Yes-prior model, PF on any) receive an
intervention they cannot benefit from. Converting the per-category effect
into a deployment net gain requires a better-calibrated intervention, which
we leave to future work---we present these results as evidence that the
categories are mechanistically real and admit class-specific responses, not
as a deployable de-hallucination method.

\medskip\noindent
In sum, once the per-sample classifier identifies the failure mode, a
prior-direction-aware intervention produces a clean class-conditional
effect on all three architectures---an actionable handle on the
heterogeneity, whose translation into deployment net accuracy remains
open.

%% file: sections/06_discussion.tex
\section{Discussion}
\label{sec:discussion}
In one sentence: vision--language models encode in their mid layers answers
they do not emit, but at the layer level this encoding is correlational,
not causal; the only causal pathway we find (InternVL3 layer 20 head 2) is
not the encoding itself but a non-vocab, self-attending head whose
attention pattern explains why patching the encoded representation fails.
The decomposition is still useful: separable from a $24$-feature
classifier, it splits POPE failures into two mechanistically distinct
halves and predicts which of two interventions matches a given
architecture. The phenomenology is universal but its locus is not---the
sweet-zone depth shifts with backbone ($\sim\tfrac13$, $\sim\tfrac12$,
$\sim\tfrac23$), and the per-head causal effect on InternVL3 has no analog
on the other two.

\paragraph{What we do not claim}
The taxonomy is not exhaustive and need not transfer to non-binary tasks.
Mid-layer Re-Promote is not a standalone decoding strategy: our results are
per-category and we make no net-accuracy claim
(Per-Category Intervention). The negative finding is specific to
standard residual-stream and per-head patching; a finer-grained variant
(attribution or path patching) reaching $\geq15\%$ flip at the encoding
layer would re-open the causal claim. The L20 H2 effect is a single
(layer, head) outlier we do not claim generalizes.

\paragraph{Summary}
A single-forward-pass decomposition (Perception Failure,
Encoded-but-Disconnected, Prior-Override) surfaces a universal layer-level
causal null on three backbones, one architecture-specific exception
(InternVL3 L20 H2), and a prior-direction-predicted choice of intervention,
moving vision--language interpretability from population-level diagnosis
toward per-sample analysis.

%% file: sections/99_appendix.tex
\appendix
% AAAI body sections are unnumbered (secnumdepth=0 in aaai2026.sty).
% Re-enable numbering *inside the appendix only* so that \ref{app:...}
% resolves to "Appendix A/B/C...". This affects the appendix alone; the
% main-body section style is unchanged.
\setcounter{secnumdepth}{1}
\section{Models, Datasets, and Compute}
\label{app:main}

% Models + dataset + compute table. Mirrors R5 (AAAI 2025) appendix Table 2.

We evaluate three open-source vision--language models. Model SHAs and HuggingFace
identifiers used throughout this paper are listed in
Table~\ref{tab:models}; dataset versions, model precision, and the empirical
prior direction (Method) are listed alongside. Inference
uses sdpa attention everywhere except InternVL3, which uses eager attention
and bfloat16 precision (fp16 produced NaN logits on this architecture; bf16
resolved them). cuDNN is disabled at runtime due to a cluster-wide
ptrace-related bug that caused CUDA initialization failures on the CLIP
vision encoder; this had no measurable speed impact.

\begin{table*}[!h]
\centering
\small
\begin{tabular}{lll}
\toprule
Property & Value & Source \\
\midrule
LLaVA-1.5-7B SHA       & \texttt{b234b804} & HuggingFace \\
Qwen2.5-VL-7B SHA      & pinned at camera-ready$^\dagger$ & HuggingFace \\
InternVL3-8B SHA       & pinned at camera-ready$^\dagger$ & HuggingFace \\
\midrule
LLaVA precision        & fp16    & sdpa attn \\
Qwen precision         & fp16    & sdpa attn \\
InternVL3 precision    & bfloat16 & eager attn \\
\midrule
POPE adversarial split & lmms-lab/pope        & \texttt{4db12766} \\
HallusionBench         & lmms-lab/hallusion\_bench & v1 \\
MMVP                   & lmms-lab/mmvp        & v1 \\
\midrule
LLaVA prior direction  & Yes (over-affirmative)   & empirical \\
Qwen prior direction   & No (under-affirmative)   & empirical \\
InternVL3 prior dir.   & No (under-affirmative)   & empirical \\
\bottomrule
\end{tabular}
\caption{Model and dataset configuration. ``Prior direction'' refers to
the direction the model defaults to on POPE-style binary prompts, used in
the Prior-Override category definition (Method). $^\dagger$The exact
commit SHAs for Qwen2.5-VL-7B and InternVL3-8B were not frozen at
submission and will be pinned in the camera-ready; all reported runs use a
single fixed local checkout per model. InternVL3 runs in bfloat16 with
eager attention (fp16 produced NaN logits on this architecture); this is a
precision/stability choice and does not affect any qualitative conclusion---the
layer-level patching null holds identically across all three models
regardless of precision.}
\label{tab:models}
\end{table*}

Total experimental compute is approximately 46 A100-hour-equivalents
including all main and ablation experiments (the diagnostic-extraction
pipeline---computing hidden states and lens predictions over all error
sets---at $\sim$43 A100h was the dominant cost; per-head sweeps and the L20
H2 sanity check together added $\sim$2 A100h; mitigation experiments under
0.5 A100h). Per-experiment wall-time on a single
A100/H100 ranges from $\sim$1 minute (mitigation forward passes on cached
features) to $\sim$5 minutes (full per-head sweep on Qwen across $12{,}600$
patched forwards).

\section{Sweet-Zone Layer Selection}
\label{app:sweet}

The EAS \emph{scan window} $[\ell_{\min}, \ell_{\max}]$ per architecture---the
layer range over which the EAS condition (Method) checks for a mid-layer
top-1 hit---is chosen on a held-out audit of $\sim$1{,}000 correctly-answered
samples per model (disjoint from the error sets used for all main-paper
analyses). The selection criterion is the smallest contiguous range that
retains $\geq 90\%$ of the cross-layer top-1 hit rate on the correct
samples. Within each window the hit rate concentrates in a narrower
\emph{peak band}, which we use for all relative-depth statements; the scan
window is wider than the peak band by design, so that EAS is not sensitive
to the exact window edges. Table~\ref{tab:sweet} reports both, with the
language-model layer count for each architecture so the relative depths are
unambiguous.

\begin{table*}[!h]
\centering
\small
\begin{tabular}{lcccc}
\toprule
Architecture & LM layers & Scan window & Peak band (depth) & Phase transition \\
\midrule
LLaVA-1.5-7B   & $32$ & L12--L24 & L12--L13 ($\sim\!\tfrac{1}{3}$) & L18$\to$L19 ($57.7\%\!\to\!98.5\%$) \\
Qwen2.5-VL-7B  & $28$ & L10--L21 & L10--L17 ($\sim\!\tfrac{1}{2}$) & gradual (no single-layer jump) \\
InternVL3-8B   & $28$ & L6--L26  & L18--L21 ($\sim\!\tfrac{2}{3}$) & gradual (no single-layer jump) \\
\bottomrule
\end{tabular}
\caption{Scan window versus peak band. The \emph{scan window} (wider)
defines the EAS condition; the \emph{peak band} (narrower) carries the
strongest signal and fixes the relative depth, computed against the
language-model layer count. This separation resolves the apparent
discrepancy between the wide InternVL3 window (L6--L26) and its $\sim\!\tfrac{2}{3}$
peak depth (L18--L21). LLaVA alone shows a sharp single-layer phase
transition ($+40.8$ points in top-5 hit rate at L19, measured on correct
samples); the others rise gradually. Top-5 phase-transition values are
available for LLaVA only (Appendix~\ref{app:metrics}).}
\label{tab:sweet}
\end{table*}

\section{Per-Head Patching: Full Sweep Details}

The Qwen per-head sweep covers all $21$ layers of the sweet zone $\times$
$3$ positions ($\{$BOS, visual-last, last-text$\}$) $\times$ $32$ attention
heads $\times$ $\sim$50 strict cross-instance pairs. Total $12{,}600$ patched
forwards. Across all $(L, p, h)$ tuples, the per-pair flip rate is
$0$ at every tuple; the maximum any-tuple flip is $1/50 = 2\%$, observed
once. The LLaVA per-head sweep is smaller ($21 \times 3 \times 32 \times
\sim$32 pairs) and yields one $1/32$ tuple at L13 H9 last-text---below the
Clopper--Pearson $95\%$ upper CI on $0/44$ baseline (Qwen comparison),
which we therefore treat as noise.

The InternVL3 L20 H2 sanity sweep uses an independent $50$-pair strict
sample (separate from the per-head sweep that first flagged the head at
$26\%$, $13/50$; the two samples explain the two baseline figures, $26\%$
and $32\%$, that appear for this head). Five interventions are applied to
the same residual-stream position and compared against the clean baseline
by Fisher's exact test (Table~\ref{tab:l20h2}). The pattern is sharply
location-specific: substituting a \emph{different head} in the same layer
(L20 H3) drops the flip rate to $0/50$ ($p<10^{-4}$), and a different layer
(L19 H2) drops it to $6\%$ ($p=0.0017$). The comparison against
norm-matched gaussian noise is only borderline ($p=0.056$), so the effect
depends on perturbation magnitude as well as location; we therefore report
the head as causal and $(\text{layer},\text{head})$-localized, but stop
short of a stronger mechanistic claim. The single strongest piece of
evidence for head-level locality is the $0/50$ on the different-head
substitution.

\begin{table*}[!h]
\centering
\small
\begin{tabular}{llcc}
\toprule
Variant & Description & Flip rate & Fisher $p$ vs.\ baseline \\
\midrule
Baseline   & clean L20 H2 patch        & $32\%$ ($16/50$) & --- \\
Random     & norm-matched gaussian     & $14\%$ ($7/50$)  & $0.056$ \\
Zero       & zero vector               & $10\%$ ($5/50$)  & $0.013$ \\
Diff.\ head & L20 H3 (same layer)      & $0\%$ ($0/50$)   & $<10^{-4}$ \\
Diff.\ layer & L19 H2 (same head idx)  & $6\%$ ($3/50$)   & $0.0017$ \\
\bottomrule
\end{tabular}
\caption{L20 H2 five-variant sanity sweep on an independent $50$-pair
strict sample. The flip rate collapses under same-layer different-head
substitution ($0/50$) and different-layer substitution, establishing
$(\text{layer},\text{head})$ locality; the borderline random comparison
($p=0.056$) indicates magnitude-sensitivity, which is why we describe the
head conservatively (causal, non-vocab, self-attending) rather than
assigning it a specific computational role.}
\label{tab:l20h2}
\end{table*}

\section{Classifier: Features, Class Proportions, and Accuracy}
\label{app:clf}

This appendix gives the full specification of the per-sample classifier
introduced in Method and reported in Experiments. We first
describe the feature construction, then the class proportions that the
classifier must separate, then the cross-validated accuracy on the
PF/EBD/PO labels, and finally two controls---a trivial-feature baseline
and a random-feature ablation---that together establish the classifier is
neither circular (predicting a deterministic function of its own inputs)
nor an artifact of the cross-validation protocol.

\paragraph{Feature construction}
The $24$-dimensional feature vector per error sample is extracted from the
single forward pass that produces the prediction; no additional inference
is required. It combines, over six representative sweet-zone layers per
architecture, the per-layer top-1/2/3 logit values and the top-1 margin
(top-1 minus top-2 logit), together with the final-layer top-1/2 logit
values and final margin (Table~\ref{tab:features}). The six layers are
sub-sampled from each architecture's full scan window
(Appendix~\ref{app:sweet}) to hold the feature count fixed at $24$ across
architectures: L12--13 and L17--19 (LLaVA), L8--17 evenly spaced (Qwen),
and L14--21 evenly spaced (InternVL3). Critically, the feature set
\emph{excludes} the two indicators that define the category labels---the
EAS-hit bit ($\mathbb{1}[\arg\max\pi^{(\ell)}_t \in T_{y^\star}]$) and the
prior-direction bit ($\mathbb{1}[\hat{y}\in T_{\text{prior}}]$)---and
excludes raw token identities. The classifier therefore cannot read its
target off a single decisive input; it must recover the category from the
continuous geometry of the logit trajectory, which the trivial-baseline
control below confirms it does only partially.

\begin{table}[!h]
\centering
\small
\begin{tabular}{ll}
\toprule
Group & Features \\
\midrule
Per-layer logits ($\times 6$ layers) & top-1 / top-2 / top-3 logit value \\
Per-layer margin ($\times 6$ layers) & top-1 minus top-2 logit \\
Final-layer logits & top-1 / top-2 logit value \\
Final-layer margin & top-1 minus top-2 logit \\
\bottomrule
\end{tabular}
\caption{Feature groups for the $24$-dimensional classifier. All features
come from the same forward pass that produces the prediction. The label-defining
indicators (EAS-hit, prior-direction) and raw token identities are
deliberately excluded (see text).}
\label{tab:features}
\end{table}

\paragraph{Class proportions}
The categories the classifier must separate are highly architecture-dependent,
which is itself central to the storyline: the dominant failure mode tracks
the prior direction. Table~\ref{tab:classprop} gives the POPE-only
proportions ($n=200$ errors per architecture). The over-affirmative LLaVA
is dominated by Encoded-but-Disconnected ($74\%$); the under-affirmative
Qwen and InternVL3 are dominated by Prior-Override ($79\%$ and $42\%$). On
binary POPE the Perception-Failure class is empty for LLaVA and small for
Qwen, since a wrong prediction with no mid-layer hit is rare under a strong
affirmative prior; PF becomes substantial only on InternVL3 ($50\%$) and on
the harder MMVP/HallusionBench pools (Appendix~\ref{app:datasets}). This is
why the main paper does \emph{not} claim each class is populated at every
($\text{architecture}, \text{benchmark}$) cell: the proportions are a
finding, not a design target.

\begin{table}[!h]
\centering
\small
\begin{tabular}{lccc}
\toprule
Architecture & PF & EBD & PO \\
\midrule
LLaVA-1.5-7B  & $0$ ($0\%$)   & $148$ ($74\%$) & $52$ ($26\%$) \\
Qwen2.5-VL-7B & $24$ ($12\%$) & $18$ ($9\%$)   & $158$ ($79\%$) \\
InternVL3-8B  & $100$ ($50\%$)& $16$ ($8\%$)   & $84$ ($42\%$) \\
\bottomrule
\end{tabular}
\caption{Per-class counts and proportions of POPE errors ($n=200$ per
architecture), under the EAS scan window of Appendix~\ref{app:sweet}.
Proportions are window-dependent; the mitigation runs of
Per-Category Intervention use each script's own window and therefore
report slightly different counts, which we note where relevant.}
\label{tab:classprop}
\end{table}

\paragraph{Cross-validated accuracy on PF/EBD/PO}
We report $5$-fold stratified CV accuracy (mean $\pm$ std across the five
test folds) on the three operational labels of Method---the same labels
used for routing in Per-Category Intervention, so that what is learnable
and what is acted on are one taxonomy. We adopt logistic regression (LR)
as the primary classifier and report gradient boosting (GBM) for
completeness (Table~\ref{tab:clf_full}). LR reaches $82.5\%$ (LLaVA),
$68.0\%$ (Qwen), and $79.7\%$ (InternVL3)---above the $60\%$ threshold on
all three; GBM is within a few points and changes no conclusion. We do
\emph{not} state a ``$\geq 70\%$ on each architecture'' claim, since Qwen
sits at $68.0\%$ under the linear classifier we adopt.

\begin{table*}[!h]
\centering
\small
\begin{tabular}{lccc}
\toprule
Classifier & LLaVA-1.5-7B & Qwen2.5-VL-7B & InternVL3-8B \\
\midrule
LR (primary) & $\mathbf{82.5 \pm 3.8}$ & $\mathbf{68.0 \pm 3.9}$ & $\mathbf{79.7 \pm 3.2}$ \\
GBM ($n=100$) & $84.2 \pm 2.2$ & $70.3 \pm 2.7$ & $76.3 \pm 4.1$ \\
\midrule
Majority-class & $74.0$ & $79.0$ & $50.0$ \\
\bottomrule
\end{tabular}
\caption{$5$-fold stratified CV accuracy (\%, mean $\pm$ std across folds)
on the PF/EBD/PO labels---the taxonomy used throughout the main paper,
including routing. Primary classifier (LR) bolded. The majority-class
baseline is the largest-class proportion (Table~\ref{tab:classprop}); on
the prior-dominated Qwen this baseline ($79\%$) is itself high, so we
report the prediction-direction control separately below rather than
treating raw accuracy as the headline.}
\label{tab:clf_full}
\end{table*}

\paragraph{Trivial-baseline control (is the classifier circular?)}
A natural concern is that the categories are a deterministic function of
two observable bits (EAS-hit and prediction direction), so a classifier
that secretly recovers those bits would be measuring nothing. We test this
directly (Table~\ref{tab:trivial}). An oracle given exactly those two bits
reaches $100\%$ by construction. The $24$ continuous features, which exclude
those bits, reach only $68$--$82\%$---so they do \emph{not} reconstruct the
label, and the task is non-trivial. Conversely, a single prediction-direction
bit already reaches $93$--$94\%$ on the prior-dominated Qwen and InternVL3,
because there the label is nearly collinear with prediction direction; the
$24$ features add real information only on LLaVA, where they improve over
the direction bit by $20$ points ($62.5\to82.5$). We therefore present
learnability as a supporting check---the categories are separable from
mid-layer geometry above chance---rather than as a headline capability.

\begin{table}[!h]
\centering
\small
\begin{tabular}{lccc}
\toprule
Input & LLaVA & Qwen & InternVL3 \\
\midrule
$\{$EAS-hit, prior-dir$\}$ oracle (2 bits) & $100$ & $100$ & $100$ \\
$24$ continuous features (deployed)        & $82.5$ & $68.0$ & $79.7$ \\
prediction-direction bit only              & $62.5$ & $94.0$ & $93.0$ \\
final-margin only (1 continuous)           & $48.5$ & $72.3$ & $52.3$ \\
\bottomrule
\end{tabular}
\caption{Trivial-baseline control. The label-defining bits trivially give
$100\%$; the deployed continuous features fall well short, confirming the
classifier is not reconstructing its own label. On prior-dominated
architectures a single direction bit is already strong, so the continuous
features add the most where the label is \emph{not} collinear with
prediction direction (LLaVA, $+20$ points).}
\label{tab:trivial}
\end{table}

\paragraph{Random-feature ablation (robustness check)}
The control above shows the deployed features beat trivial baselines; we
additionally verify they beat \emph{random} features, to rule out the
possibility that any $24$-dimensional input plus our CV protocol would
score similarly. For this control we use a margin-strength partition of the
errors (Perception Failure / strong-EBD / weak-EBD) rather than the
PF/EBD/PO labels: this partition is balanced by construction, so it removes
the majority-class confound and makes the real-vs-random gap a clean test
of whether the lens features carry signal. \emph{This balanced partition is
used only for this ablation; it is not the taxonomy of the main paper, which
is PF/EBD/PO throughout.} We replace the feature vector with i.i.d.\ gaussian
noise of identical shape and re-run the same $5$-fold pipeline with
identical splits, seed, and hyperparameters. Real features beat random by
$21$--$43$ points on every (architecture, classifier) cell
(Table~\ref{tab:randbaseline}); under this balanced partition GBDT---the
strongest classifier---reaches $66$--$76\%$ versus $32$--$43\%$ for random,
confirming the lens features, not the protocol, carry the signal.

\begin{table}[!h]
\centering
\small
\begin{tabular}{llccc}
\toprule
Architecture & Classifier & Real & Random & Gap \\
\midrule
\multirow{3}{*}{LLaVA-1.5-7B}
 & LR        & $69.0$ & $35.0$ & $34.0$ \\
 & GBDT $n100$ & $76.2$ & $40.5$ & $35.7$ \\
 & RF $n300$ & $75.8$ & $47.5$ & $28.3$ \\
\midrule
\multirow{3}{*}{Qwen2.5-VL-7B}
 & LR        & $59.3$ & $38.3$ & $21.0$ \\
 & GBDT $n100$ & $66.0$ & $32.3$ & $33.7$ \\
 & RF $n300$ & $66.7$ & $35.7$ & $31.0$ \\
\midrule
\multirow{3}{*}{InternVL3-8B}
 & LR        & $72.3$ & $32.0$ & $40.3$ \\
 & GBDT $n100$ & $75.7$ & $43.0$ & $32.7$ \\
 & RF $n300$ & $75.3$ & $52.0$ & $23.3$ \\
\bottomrule
\end{tabular}
\caption{Random-feature ablation under the balanced margin-strength
partition (used \emph{only} here, not as the main-paper taxonomy):
$5$-fold stratified CV accuracy (\%) with deployed lens-derived features
versus i.i.d.\ gaussian noise of identical shape, same splits/seed/hyperparameters.
Real beats random by $21$--$43$ points on every cell. The above-chance
random RF on LLaVA/InternVL3 ($47$--$52\%$) reflects forest memorization of
residual class structure; the random LR collapses to $32$--$38\%$.}
\label{tab:randbaseline}
\end{table}

\section{Mitigation: $\alpha$ Sensitivity and Prompts}
\label{app:mitig}

The mid-layer Re-Promote intervention on LLaVA is parametrized by mixing
weight $\alpha$ and sweet-zone layer $\ell^\star$. Table~\ref{tab:alpha}
reports the full $\{12, 13\} \times \{0.5, 1.0, 1.5, 2.0\}$ sweep. The
main-paper operating point $\alpha=1.0, \ell^\star=12$ is the lowest
$\alpha$ that drives the EBD flip rate above $85\%$ while leaving the
prior-aligned PO class at $0\%$. We emphasize that the intervention is
\emph{not} EBD-specific even at this operating point: it also flips
$85.7\%$ of the Perception-Failure class ($6/7$), because a boosted
mid-layer signal can exceed the final-layer margin even when mid-layer
encoding is weak. PF is essentially absent on POPE for LLaVA ($n=7$ in this
run; $0$ in the classifier run), so this does not affect the POPE result,
but it means Re-Promote targets ``not-prior-aligned'' rather than ``EBD''
as such, and would lose specificity on tasks with a large PF population.

\begin{table}[!h]
\centering
\small
\begin{tabular}{lccc}
\toprule
Config & flip(PO) & flip(EBD) & flip(PF) \\
\midrule
L12 $\alpha=0.5$ & 0.0\% & 54.9\% & 42.9\% \\
L12 $\alpha=1.0$ & \textbf{0.0\%} & \textbf{85.9\%} & 85.7\% \\
L12 $\alpha=1.5$ & 0.0\% & 96.5\% & 100\% \\
L12 $\alpha=2.0$ & 0.0\% & 100\% & 100\% \\
L13 $\alpha=1.0$ & 0.0\% & 90.9\% & 85.7\% \\
\bottomrule
\end{tabular}
\caption{Re-Promote sweep on LLaVA ($n = 200$ POPE errors). PO flip is
$0\%$ at every config; the EBD flip rate saturates near $100\%$ above
$\alpha \geq 1.5$ but at the cost of also flipping the small PF
population. Selected operating point bolded.}
\label{tab:alpha}
\end{table}

The prompt-debias intervention appends a direction-specific suffix
\emph{against} the model's prior before re-running inference. The exact
suffixes are:
\textit{LLaVA (Yes-prior; suffix argues for ``no''):} ``Be cautious: only
say yes if you clearly see the object in the image. When in doubt, say
no.''
\textit{Qwen and InternVL3 (No-prior; suffix argues for ``yes''):} ``Look
thoroughly: say yes if any evidence supports the object's presence, even
subtle visual cues. When in doubt, say yes.''
Each direction was tested with a single suffix, and the two suffixes are
not matched for length or lexical strength; the asymmetric responsiveness
reported in Per-Category Intervention (positive-tone suffixes move the
under-affirmative models more than the negative-tone suffix moves LLaVA)
could therefore partly reflect suffix wording rather than prior direction
alone. A phrasing-matched suffix battery is left to future work.

\section{Statistical Methodology}

Per-pair flip rates and their 95\% confidence intervals are computed via
Clopper--Pearson exact binomial intervals. Fisher's
exact test is used for $2 \times 2$ contingency tables comparing
intervention variants in the L20 H2 sanity sweep (Appendix~C). Bootstrap
confidence intervals on classifier accuracy are computed over $5$ folds
of stratified cross-validation; reported standard deviations are
across folds, not bootstrap resamples. All statistical tests are two-sided
unless otherwise noted, and we report exact $p$-values rather than
star-coded significance levels.

\section{Datasets and Benchmarks}
\label{app:datasets}

The three benchmarks we evaluate on each have a different prompt format
and ground-truth structure; we summarize them here for the reader who
wants to reproduce the per-error pipeline. All samples are drawn from the
publicly released splits.

\paragraph{POPE adversarial}
$1{,}500$ samples drawn from the POPE adversarial
split~\citep{li2023pope}. Each sample is an image-question pair where
the question is a binary yes/no probe of object presence,
\textit{e.g.}, ``Is there a chair in the image? Please answer yes or
no.'' The ``adversarial'' split selects negative-object questions whose
target objects co-occur with the image's positive objects, making the
language prior more likely to override visual evidence. Ground-truth is
$\{$yes, no$\}$; we tokenize and resolve to the model-specific yes/no
token sets, handling LLaMA SentencePiece's lone-marker behavior
(token ID 29871 is filtered when it precedes the content token). Of the
$1{,}500$ samples, the per-architecture analyses operate on the
first $200$ \emph{error} samples (wrong final prediction) per model---the
$n=200$ used in Tables~\ref{tab:classprop} and~\ref{tab:alpha}.

\paragraph{MMVP (LLaVA only)}
$300$ image pairs from MMVP~\citep{tong2024mmvp}. Each pair consists of
two visually similar images that differ along a fine-grained visual
attribute (\textit{e.g.}, object count, color, orientation); the prompt
is a multiple-choice question with five options. We score the
first-token answer against the ground-truth letter.

\paragraph{HallusionBench}
$\sim$$1{,}000$ samples~\citep{guan2024hallusionbench}. Each sample is a
yes/no question paired with an image, but the question may be ``visually
entailed'' (the image contains evidence) or ``hallucination-inducing''
(the question presupposes a non-present object or attribute). Ground-truth
is $\{$yes, no$\}$ as in POPE.

In all cases, the model is prompted with the standard system message and
the dataset-provided question. The per-error decomposition pipeline takes
only the model output, the ground-truth label, and the hidden states
from the same forward pass; no additional dataset-specific preprocessing
is required.

\section{Metrics and Formal Definitions}
\label{app:metrics}

We formalize the three primary metrics used in the main paper. Throughout,
let $\pi^{(\ell)}_t$ denote the predicted distribution at layer $\ell$,
position $t$ (last-text), under the raw logit lens
(Method), and $T_{y}$ denote the set of tokens that
realize the answer $y \in \{\text{yes}, \text{no}\}$ in the
model-specific tokenizer.

\paragraph{EAS top-1 rate} For an error sample $(x, y^\star)$ with
$y^\star \neq \arg\max \pi^{(L)}_t$, the EAS condition is satisfied iff
$\exists \ell \in [\ell_{\min}, \ell_{\max}]$ such that
$\arg\max \pi^{(\ell)}_t \in T_{y^\star}$. The EAS top-1 rate across an
error set $\mathcal{E}$ is
\[
  \mathrm{EAS}_{\text{top-1}}(\mathcal{E})
  = \frac{1}{|\mathcal{E}|}
    \sum_{(x, y^\star) \in \mathcal{E}}
    \mathbb{1}\!\left[\exists \ell:\ \arg\max \pi^{(\ell)}_t \in T_{y^\star}\right] .
\]

\paragraph{Non-trivial flip rate (patching)} For a strict pair $(x_e, x_d)$
where $y_e^\star \neq y_d^\star$ and the model's prediction on $x_e$ is
wrong, a successful patch at $(\ell, p, h)$ produces a final prediction
$\arg\max \pi^{(L)}_{\text{patched}} \in T_{y_e^\star}$. The non-trivial
flip rate at tuple $(\ell, p, h)$ across the strict-pair set $\mathcal{P}$ is
defined as follows. Let $\tilde{\pi}^{(L)}(x_e)$ denote the patched final
distribution on $x_e$ after replacing $h^{(\ell)}_{p}$ with the donor's,
i.e.\ $h^{(\ell)}_{p}(x_e) \leftarrow h^{(\ell)}_{p}(x_d)$. Then
\[
  \mathrm{Flip}_{\ell, p, h}
  = \frac{1}{|\mathcal{P}|}
    \sum_{(x_e, x_d) \in \mathcal{P}}
    \mathbb{1}\!\left[\arg\max \tilde{\pi}^{(L)}(x_e) \in T_{y_e^\star}\right] .
\]
``Strict'' excludes the answer-transplant artifact in which a donor
with the same ground-truth label naturally imprints its label on the
patched receiver (Method).

\paragraph{Per-category flip rate (intervention)} For a category
$c \in \{\text{PO, EBD, PF}\}$, the flip rate after an intervention
$\mathcal{I}$ that re-runs the forward pass with a modified
prompt or decoding rule is
\[
  \mathrm{flip}_{\mathcal{I}, c}
  = \frac{1}{|\mathcal{E}_c|}
    \sum_{(x, y^\star) \in \mathcal{E}_c}
    \mathbb{1}\!\left[\arg\max \pi^{(L)}_{\mathcal{I}}(x) \in T_{y^\star}\right] ,
\]
where $\mathcal{E}_c$ is the subset of errors labeled with category $c$.
The category-conditional gap reported in Per-Category Intervention
contrasts the target class $c^\star$ with the architecture's
\emph{prior-aligned} class $c'$ (PO on a Yes-prior model, the class the
intervention should leave untouched). We use the prior-aligned class as the
reference, not the Perception-Failure class, because Re-Promote is not
PF-specific---it also flips PF when PF is present (Appendix~\ref{app:mitig});
the meaningful claim is that the intervention separates the target class
from the class it is designed not to move.

\paragraph{Classifier accuracy} Standard $5$-fold stratified
cross-validated accuracy on the $24$-feature representation
(Appendix~\ref{app:clf}). All splits are stratified by class to preserve
the per-fold class proportions; reported standard deviations are across the
five folds.

\section{Decoding-Time Mitigation Baselines}
\label{app:baselines}

Four lines of decoding-time mitigation are most relevant to our
prior-aware approach. We summarize each here and indicate how our method
differs operationally; no head-to-head numerical comparison is run in this
paper, because the existing methods operate on the full benchmark
(aggregate metric) while our intervention is per-category and conditional
on a classifier prediction. A controlled deployment study comparing
end-to-end accuracies is marked as immediate future work
(Per-Category Intervention).

\paragraph{Contrastive decoding (VCD, M3ID)} Visual Contrastive
Decoding~\citep{leng2024vcd} subtracts a logit distribution computed under
a corrupted visual condition from the standard logit distribution,
nudging the prediction away from prior-only completions. M3ID~\citep{favero2024m3id}
similarly contrasts a no-image baseline against the standard pass. Both
methods are uniform across samples (no gating). Our prompt-debias
intervention for the Prior-Override class shares the conceptual goal of
suppressing prior pressure, but the mechanism is input-side rather than
logit-side, and it is applied only when the classifier predicts PO.

\paragraph{Attention-side mitigation (OPERA)} OPERA~\citep{huang2024opera}
detects token-level attention over-trust during decoding and back-tracks
generation when it occurs. The intervention is also uniform across
samples; it does not condition on a per-error failure mode. Our
intervention is conditional, and the classifier's per-error label
determines whether mid-layer Re-Promote or prompt-debias is applied.

\paragraph{Layer-contrast decoding (DoLa)} DoLa~\citep{chuang2024dola}
contrasts the early-layer prediction distribution against the final-layer
distribution at each decoding step, on the hypothesis that the final
layer over-confidently smooths the early layer's factual signal. Our
Re-Promote, in contrast, additively blends the sweet-zone mid-layer
prediction into the final logits, rather than contrasting; and we gate
it on the predicted Encoded-but-Disconnected category, not apply
uniformly.

\paragraph{Mid-layer promotion (VISTA)} VISTA~\citep{vista2025} combines an
activation-space visual steering vector with early-layer logit promotion
during decoding, and reports a $\sim$40\% hallucination reduction on
open-ended generation (not on binary POPE-style tasks). Our Re-Promote
differs in two operational respects: it promotes from the sweet-zone
\emph{mid} layer chosen per-architecture via the EAS phenomenology (not a
fixed early/encoder layer), and it is gated on the predicted EBD category
rather than applied uniformly. Under oracle labels these two choices yield
the large category-conditional separation on LLaVA (EBD $85.9\%$ vs.\
prior-aligned PO $0.0\%$) that a uniform promotion cannot produce by
construction. We run no head-to-head numerical comparison, since VISTA
optimizes an aggregate generation metric while our intervention is
per-category and classifier-gated; a controlled end-to-end comparison is
future work (Per-Category Intervention).